\documentclass[wcp]{jmlr}

\usepackage{color}
\usepackage{algorithm}
\usepackage{algpseudocode}
\usepackage[T1]{fontenc}
\usepackage{graphics}

\title[Random Gradient]{Fast, Better Training Trick --- \\Random Gradient}

  \author{\Name{JK, Wei} \Email{16408400236@stu.hut.edu.cn}\\
  \addr Hunan University of Technology
 }

\begin{document}

\maketitle

\begin{abstract}
	In this paper, we will show an unprecedented method to accelerate training and improve performance, which called random gradient (RG). This method can be easier to the training of any model without extra calculation cost, we use Image classification, Semantic segmentation, and GANs to confirm this method can improve speed which is training model in computer vision. The central idea is using the loss multiplied by a random number to random reduce the back-propagation gradient. We can use this method to produce a better result in Pascal VOC, Cifar, Cityscapes datasets.
\end{abstract}

\section{Introduction}
When deep learning shows excellent results in more and more areas, it is becoming more and more important to understand its internal working principles and can explain some phenomena. Back-propagation~\cite{Authors12} plays an important role in deep learning field, the principle is use of gradients calculated for each iteration to update the parameters. It build the foundation for alexnet's excellent performance in ImageNet~\cite{Authors19} competition in 2012, though there are still many problems in backpropagation, such as gradient vanishing/exploding, etc, so far it is still an open question. One of the reason is that the derivative in the nonlinear layer may tend to very small or very big value, which is exacerbated by the accumulation of multiple layers, resulting in the network cannot become deeper or hard to train. Researchers mostly solved this problem by using ReLU~\cite{Authors26} activation functions, better initialization methods~\cite{Authors21}, and skip connection by ResNet~\cite{Authors17}, which is proposed by Kaiming He. It either can solve this problem in terms of network structure and can be quickly transmitted through the residual structure effectively.

In this paper, we multiply the loss of loss function calculation by a random number between 0 and 1, and use it as a new loss for parameter optimization, in other words, the new loss is less than original loss because of a random number. Of course it can also be called random loss, however, since the derivation process will indirectly result in a random gradient, we are collectively called as random gradient. We used a variety of learning rates and momentum to experiment random gradient methods, drew some theory based on experimental phenomena, obtained the characteristics of the model at random gradients and some training skills. Experiments show that this is faster and better result, we have the following contributions:

\begin{itemize}
	\item We designed a random gradient for backpropagation, multiplying the loss value with a random number.
	\item Experiments show that the random gradient can effectively accelerate the convergence and reduce oscillation during optimization.
	\item We draw a close connection between RG, learning rate, task categories, and momentum, and come to a complete set of theory.
\end{itemize}

We have not shown the results on some large datasets, the main reason is that we cannot afford the time such as ImageNet~\cite{Authors19} or MSCOCO~\cite{Authors27}. The proposed method is evaluated on the Pascal VOC 2012 datasets~\cite{Authors16} for semantic segmentation, and cifar datasets~\cite{Authors23} for image classification task, also have cityscapes~\cite{Authors47}, maps (scraped from Google Maps.) dataset for GANs, these experiments can be approximated as the performance of the network at different scales and different tasks, we also hope that if researchers are interested in this work, they can further study and improve performance. The following is a brief introduction to semantic segmentation, image classification, and generation adversarial networks.

Semantic segmentation task contains 20 foreground object classes and one background class, dataset contains 1,464 (train), 1,449 (val), and 1,456 (test) pixel-level annotated images. The performance is measured in terms of pixel intersection-over-union averaged across the 21 classes (mIOU), but the commonly used extra annotations datasets~\cite{Authors48} will not be used to improve accuracy. Inspired by~\cite{Authors48}, we use the "poly" learning rate policy that the current learning rate equals to the base one multiplying $(1-\frac{iter}{max-iter})^{power}$. We set the base power to 0.9, we use the random mirror for data augmentation.

Image classification task which using several excellent networks for training. Two CIFAR datasets \cite{Authors23} consist of colored natural images with 32$\times$32 pixels, CIFAR-10 consists of images drawn from 10 and CIFAR-100 from 100 classes. The training and test sets contain 50,000 and 10,000 images respectively. We adopt a standard data augmentation scheme (mirroring/shifting) that is widely used for these two datasets. For preprocessing, we normalize the data using the channel means and standard deviations.

Generative Adversarial Networks (GANs)~\cite{Authors40,Authors41} have achieved impressive results in image generation~\cite{Authors42,Authors43}, and representation learning~\cite{Authors44}. The key to GANs’ success is the idea of an adversarial loss that forces the generated images to be, in principle, indistinguishable from real images. This is particularly powerful for image generation tasks, as this is exactly the objective that much of computer graphics aims to optimize. We used the excellent pix2pix~\cite{Authors46} network to experiment, batch size is 8, and other settings were strictly followed by the article.
\begin{figure}[t]
	\centering
	\includegraphics[width=12.5cm]{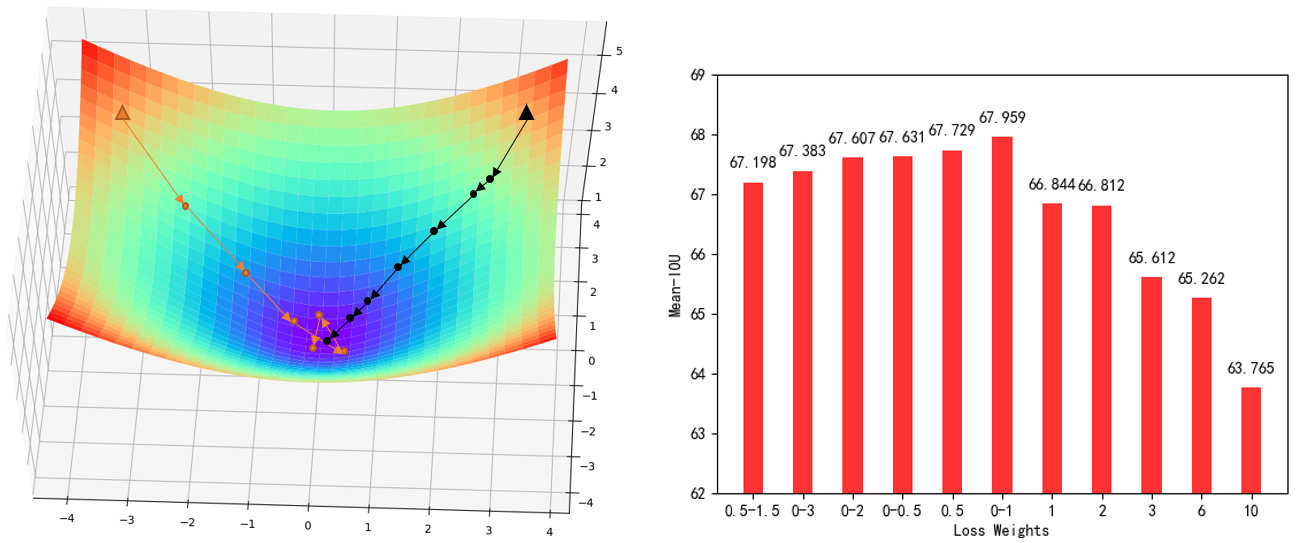}
	\caption{{\bf Left}: an example of the loss function topology. {\bf Right}: When using PSPNet on semantic segmentation results, "-" represents the upper and lower limits of the random number. we set the "batchsize" to 4 during training on a GTX1060 and "cropsize" is 224 $\times$ 224.}
	\label{fig:1}
\end{figure}

\section{Background and Inspiration}
In the area of accelerating training, researchers have put more attention on how to adjust the hyperparameters such as learning rate, batch size and momentum or design an optimization algorithm to improve performance, both them exists extensive literature on accelerating training. In this paper, convergence stop as a criterion for judging whether training is completed.

In~\cite{Authors02,Authors03,Authors07,Authors09}, researchers get a way to improve speed of convergence by changing the learning rate or adjustment the batch size. In \cite{Authors06}, the authors analyzed in detail three factors influencing minima, which is learning rate, batch size and the variance of the loss gradients, and experimentally verify that the noise $n = \frac{\eta}{S}$ ($\eta$ is the learning rate, $S$ is the batch size) determines the width and height of the minima towards which SGD converges. In~\cite{Authors01}, the authors found a super convergence phenomenon, one of the key elements of super convergence is training with cyclical learning rates~\cite{Authors07} and a large maximum learning rate.

Another direction is design a new optimization algorithm to achieve improvement, such as~\cite{Authors04,Authors05}, they based on adaptive estimates of lower-order moments, generally faster than SGD convergence, however, in some cases it is inferior to SGD performance. We also tested the performance of the random gradient method on Adam and proved that our method is not limited to SGD.

\cite{Authors13} proposes that the information transmitted in opposite direction in the brain is an order of magnitude larger than that passed forward, and neuroanatomy also confirms this view, this book gives us a deep understanding of the brain feedback mechanism. Although the current artificial neural network is very different from the brain, we wish to do some similar experiments. So we tested a variety of weights applied to the loss function, to simulate the feedback mechanism of the brain. On the right side of Figure \ref{fig:1} is the result which measured using PSPNet-101~\cite{Authors37} under the same conditions, since we have fixed the learning rate and momentum, so we can directly see the influence of the RG on the model. What surprised us is that when the weight becomes larger, the result will be worse, but when the weight is random reduced, the effect will be better, when the weight becomes a random number of zero to one, the effect is best, and there is also the same effect on image classification.

\begin{figure}[t]
	\centering
	\includegraphics[width=12.5cm]{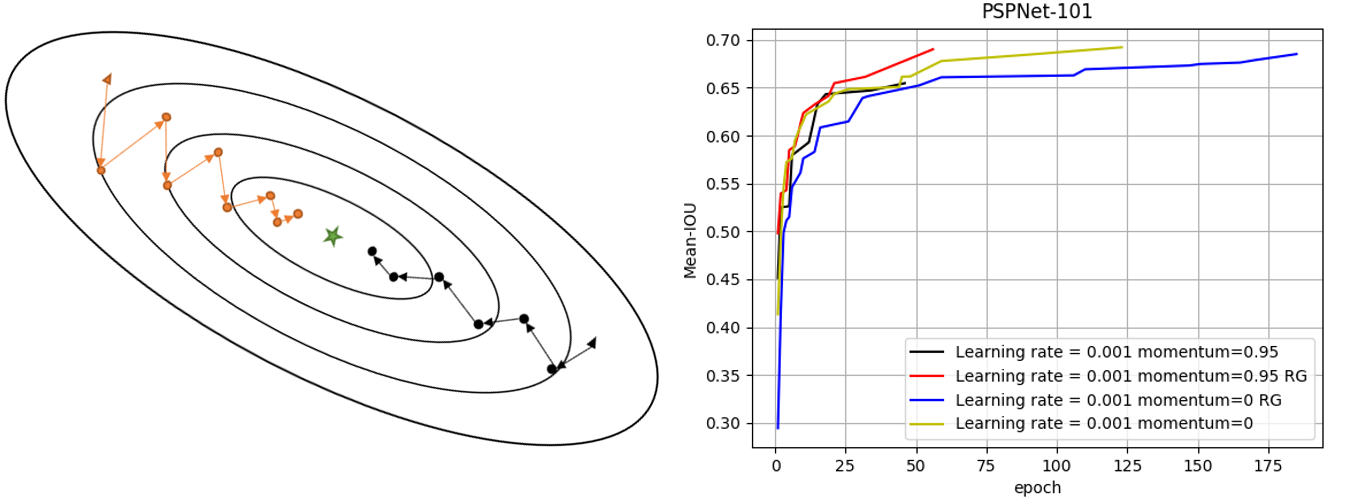}
	\caption{{\bf Left}: Hessian matrix with ill-conditioning. {\bf Right}: Comparison between random gradient and momentum, RG represents random gradient.}
	\label{fig:1_2}
\end{figure}

The left side of Figure \ref{fig:1} can be more intuitive understanding, suppose we only have two parameters that need to be optimized, this can be visualized in a three-dimensional coordinate system, the vertical axis is the value of the loss function, our finial goal is to minimize the loss function. We selected stochastic gradient descent method (SGD) for parameter updating, but the adjustment of the learning rate in SGD is crucial, excessively large may result in missing the local minimum or a better solution; too small may cause the parameter updating too slow. The orange line shows the gradient descent under normal conditions, it can be seen that there is a great deal of oscillating near the optimal value, usually we need to constantly adjust the learning rate. The black line on the right is the method proposed by us. Use the loss to multiply by a random number, it will ensure that there will not much oscillate in the optimization process under the same conditions, and we use the momentum~\cite{Authors28} method to make up for the lack of gradients, it can actually be faster convergence to local minimum or a better solution. Assume that in a more complicated example, such like the left side of Figure \ref{fig:1_2}, gradient descent direction may oscillate during optimization, random gradient method can effectively reduce this phenomenon by randomly decreasing the updated gradient each time.
\section{Analysis Optimization Method}
Gradient descent is an optimization method that uses the slope as computed by the derivative to move in the direction of greatest negative gradient to iteratively update a variable. That is, given an initial point {$x_0$}, gradient descent proposes the next point to be:
\begin{equation}
x = x_0 - \eta\frac{dJ}{dx_0}
\end{equation}

When $\eta$ is the learning rate, fraction is the derivative of the loss function to $x_0$. It can be seen that our method random reduce the value on the right meanwhile the learning rate remains unchanged.

Stochastic gradient descent is currently the most widely used optimization method, there are substantial discussion that why this solutions generalize so well in grateful literature~\cite{Authors08,Authors10,Authors11}. At each step $t$, a mini batch of $B$ samples $x_i$ is selected from the training set, the gradients of loss function $\nabla L(x_i,w)$ are computed from this subset, and networks weights $w$ are updated based on this stochastic gradient descent:
\begin{equation}
w_{t+1} = w_{t} - \eta\frac{1}{B}\sum_{i=1}^{B}\nabla L(x_i,w)
\end{equation}

Recently, many researchers no longer use vanilla SGD, instead preferring SGD with momentum~\cite{Authors28}. Momentum-based stochastic gradient descent methods are widely used in practice for training deep networks:
\begin{equation}
\begin{aligned}
&v = \alpha v - \eta\frac{1}{B}\sum_{i=1}^{B}\nabla L(x_i,w)\\
&w_{t+1} = w_{t} + v
\end{aligned}
\label{equation_3}
\end{equation}

It can be seen that the momentum method add the influence of the previous gradient on the present, in~\cite{Authors09}, the authors believes increasing the momentum coefficient will accelerate convergence, though it is likely to lose some accuracy. And we found that the random gradient method is dependent on momentum to compensate for the loss of random numbers against gradients, this is also the reason why rapid convergence can be achieved even when the gradient decreases. Supposing that the random number is very small, and the new gradient is very small too, when the momentum method is not to be used, the parameter update speed will be very slow. At this moment, the momentum method ensures that even if the current gradient is very small, the parameters can be updated greatly. It sounds very reasonable, but in the experiment, momentum only works under certain conditions. Like in classification tasks, our highest point is only 0.5, but in semantic segmentation tasks, we generally adjust it to 0.9 as the lowest point. We believe that the main reason is the requirement about the learning rates of two tasks are different. It is well known that classification tasks require greater learning rates. Under the assumption above, the gradient update is mainly from the current gradient in the classification task, and the gradient update is mainly from the accumulation of the previous gradient in the segmentation task. As can be seen in the right side of Figure \ref{fig:1_2}, momentum plays a crucial role in the speed of convergence. Compared to the normal method, the random gradient method hadn't lose precision under the precondition of accelerating convergence.

\begin{figure}
	\centering
	\includegraphics[width=12.5cm]{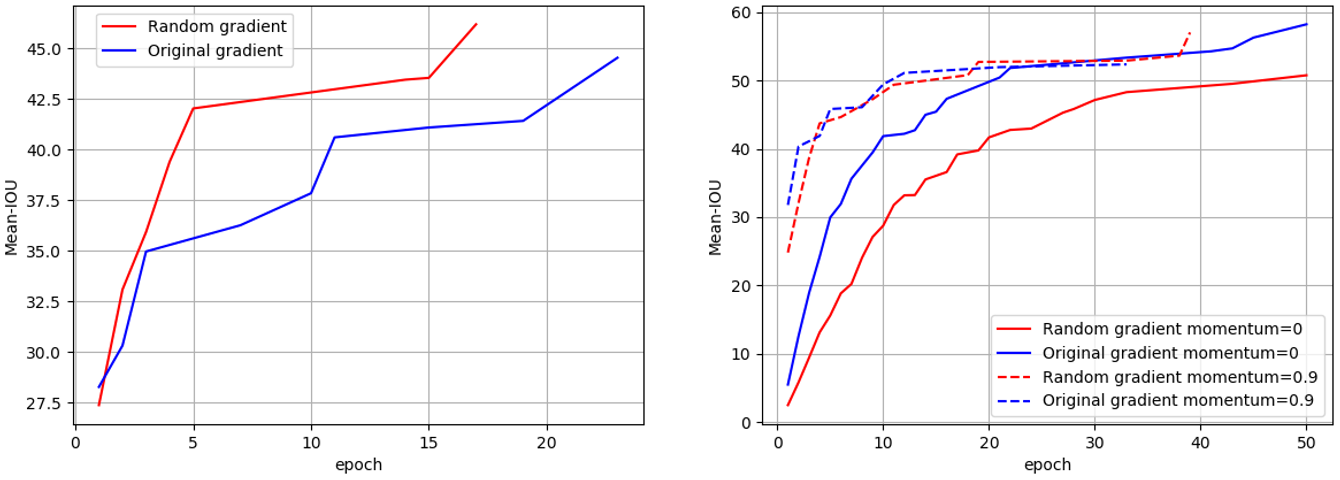}
	\caption{This experiment was conducted using ResNet101-DUC+HDC at a learning rate of 0.001 using the SGD optimization algorithm on the semantic segmentation tasks. {\bf Left}: Comparison between random gradient and original gradient when momentum = 0.95. {\bf Right}: Comparison between random gradient and original gradient when momentum is 0 and 0.9.}
	\label{fig:2}
\end{figure}

As can be seen in the Figure \ref{fig:2}, we use ResNet-DUC+HDC~\cite{Authors49} to experiment. When the momentum is 0, the efficiency of random gradient method is not the highest, and it not obvious improve the convergence speed, when the momentum is 0.9, the advantage of the random gradient is obvious, the accuracy is better than the original gradient, and compared to using the original gradient method with momentum=0, the accuracy has not been lost. On the left side of Figure \ref{fig:2}, when the momentum of the model is 0.95, if the comparison under the same conditions, convergence speed and accuracy have great improvement, but it will inevitably lose some accuracy.

We can further hypothesis that when the loss weight increases, reducing the learning rate can theoretically achieve a certain increase, such is the fact, the bottom right of Figure \ref{fig:3} is the result of this experiment. However, the consequence of this is that the convergence speed becomes slower and more complicated, this is not what we want, we do not have more in-depth experiments.

In addition, regularization~\cite{Authors25} is also often used on optimize models, such as dropout~\cite{Authors50} (This article does not explain the dropout in detail), others like $L_{1}$ regularization and $L_{2}$ regularization (also called weight decay), make the parameter as close as possible to zero or direct equal to zero. As report by \cite{Authors38}, $L_{2}$ regularization sometimes even helps optimization. In the experiment, we used $L_{2}$ regularization = 0.0005. The basic formula is as follows:
\begin{equation}
w_{t+1} = w_{t} - \eta(\frac{1}{B}\sum_{i=1}^{B}\nabla L(x_i,w) + L_{2}w)
\end{equation}

For convenience, we denote the parameters in a network as $\theta\in R^{N}$ and $f(x)$ is the loss function, $\sigma$ is non-linearity layers, $x$ is input, a network can be simplified as:
\begin{equation}
f(x) = \sigma(w_{l}\sigma(w_{l-1}\sigma(w_{l-2}\dots\sigma(w_{0}x + b_0)\dots) + b_{l-1}) + b_l)
\end{equation}

The Hessian-free optimization method was proposed by~\cite{Authors31} suggests a second order solution that utilizes the slope information contained in the second derivative (i.e., the derivative of the gradient $\nabla_{x}f(x)$), the main idea of the second order Newton's method is that the loss function can be locally approximated by the quadratic as:
\begin{equation}
f(x) \approx f(x_{0}) + (x-x_{0})^{\top}\nabla_{x}f(x_{0}) + \frac{1}{2}(x-x_{0})^{\top}H(x-x_{0})
\end{equation}

Where $H$ is the Hessian, or the second derivative matrix of $f(x_{0})$. In general, it is not feasible to compute the Hessian matrix, which has $\Omega(N^2)$ elements, where $N$ is the number of parameters in the network, but it is unnecessary to compute the full Hessian. The Hessian expresses the curvature in all directions in a high dimensional space, but the only relevant curvature direction is in the direction of steepest descent that SGD will traverse. This concept is contained within Hessian-free optimization, as \cite{Authors31} suggests a finite difference approach for obtaining an estimate of the Hessian from two gradients:
\begin{equation}
H(x) = \lim_{\delta \to 0}\frac{\nabla f(x+\delta)-\nabla f(x)}{\delta}
\end{equation}

Where $\delta$ should be in the direction of the steepest descent. Although Hessian-free optimization method has not been widely used due to its impractical to invert or even store the Hessian matrix and promotion effect is not obvious, but we still consider it is necessary to mention the Hessian-free matrix to help us to understand model optimization more deeply in second-order optimization method.

At the end of this section, we will mention the current optimization method Adam, which can converge more quickly. Adam~\cite{Authors04} is a simple and computationally efficient algorithm for gradient-based optimization of stochastic objective functions, but it takes extra memory and computing resources. We have further verified the effectiveness in this paper, it is proved that our method is also suitable for this kind of optimization algorithm.

\section{Random Gradient}
The above shows the basic formulas of optimization methods, it can be seen that the size of the learning rate and gradient directly determines the extent of the update. However, because the learning rate is associated with the batch size and is limited by the memory size of the hardware, our main breakthrough is to adjust the gradient. RG is available in almost all machine learning frameworks, such as mxnet~\cite{Authors15}, pytorch~\cite{Authors14}. Furthermore, our approach theoretically can be applied to nearly every existing deep learning architecture. The basic code structure is as follows:
\begin{verbatim}
for input, target in dataset:
    optimizer.zero_grad()
    output = model(input)
    loss = criterion(output, target)
    loss_random = loss * random()
    loss_random.backward()
    optimizer.step()
\end{verbatim}

Using RG can speed up training, although theoretically every parameter update gradient is attenuated, our experiment still have achieved great success. Similar to what is shown in Equation \ref{equation_3}, momentum method uses the previous gradient to correct the problem of the current gradient, the RG by randomly reducing the current gradient to improves performance, and it can be combined with the momentum method, make up for the impact of RG on the gradient.
\begin{table}[!htp]
	\small
	\begin{center}
		\begin{tabular}{l|l|l|l|l|l|l|l|l}
			\hline
			Learning Rate & 0.01 & 0.01 & 0.001 & 0.001 & 0.01 & 0.01 & 0.0001 & 0.0001  \\
			\hline
			Random Gradient &  &  $\surd$  &   &  $\surd$  &  &  $\surd$  &   &  $\surd$  \\
			\hline
			Momentum & 0.95 & 0.95 & 0.95 & 0.95 & 0.90 & 0.90 & 0.95 & 0.95 \\
			\hline
			Mean IOU & 31.940\% & 28.357\% & 65.473\% & 68.014\% & 28.121\% & 47.620\% & 69.076\% & 68.681\% \\
			\hline
		\end{tabular}
	\end{center}
	\caption{This experiment uses PSPNet101 sets the batchsize is 4 and the cropsize is 224$\times$224, running on a single GTX1060.
		\label{table1}
	}
\end{table}
\subsection{Theoretical Analysis}\label{Theoretical Analysis}
The most common training method is when the model stops converge is to decay the learning rate, which factor is typically 0.1. When decay two or three times, the model becomes unable to converge through the decay learning rate, normally, this means that training can stop. But in this paper, using random gradient strategy, the model can converge under a smaller gradient, and to explain random gradient can start with the second-order Taylor series expansion of the cost function:
\begin{equation}
f(x) \approx f(x_{0}) - \eta g^{\top}g + \frac{1}{2}\eta^{2}g^{\top}Hg
\end{equation}
\cite{Author52} states: \textit{There are three terms here: the original value of the function, the expected improvement due to the slope of the function, and the correction we must apply to account for the curvature of the function. In many cases, the gradient norm does not shrink significantly throughout learning, but the $g^{\top}Hg$ term grows by more than an order of magnitude.}

When the model has been oscillating without any performance improvement, it can be considered that $g^{\top}Hg$ is already large enough to affect convergence. The result is that learning becomes very slow despite the presence of a strong gradient, and the model will continue to oscillate. \textbf{The random gradient strategy has come to the fore, on the premise of no loss of convergence speed, smaller gradient becomes the key sir, scored twice to further improve performance.}


\begin{figure}[!h]
	\centering
	\includegraphics[width=12.5cm]{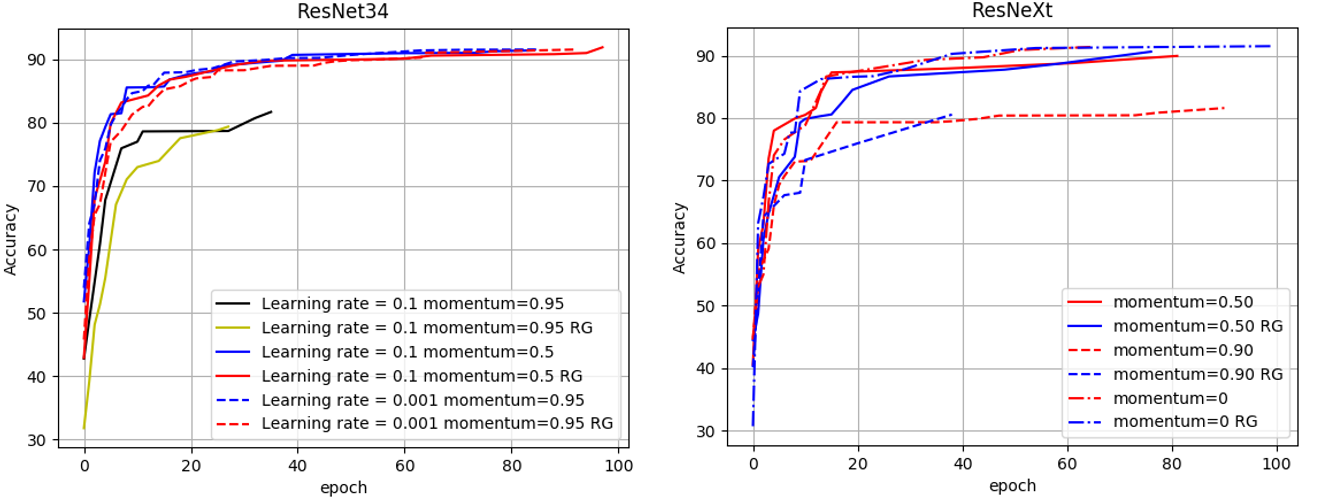}
	\caption{{\bf Left}: Comparison between random gradient and original method When changing Learning rate or Momentum. {\bf Right}: The Effect of Momentum on the ResNeXt.}
	\label{fig:7}
\end{figure}
\section{Experiment and Analysis}
In this section, we will present some experimental data and analysis. However, it is frustrating that we currently have only one GTX1060, it cannot support excessive data calculations, perhaps the experimental results did not reach the highest level but fair and convincing experiments will prove the above results.

What needs to be clarified is, although the random gradient method is very simple, but because we are involved in three areas and a variety of experiments, the experiment code will be published in \url{https://github.com/leemathew1998/RG} to facilitate researchers to obtain some details not mentioned in this paper.
\begin{figure}[h]
	\centering
	\includegraphics[width=12.5cm]{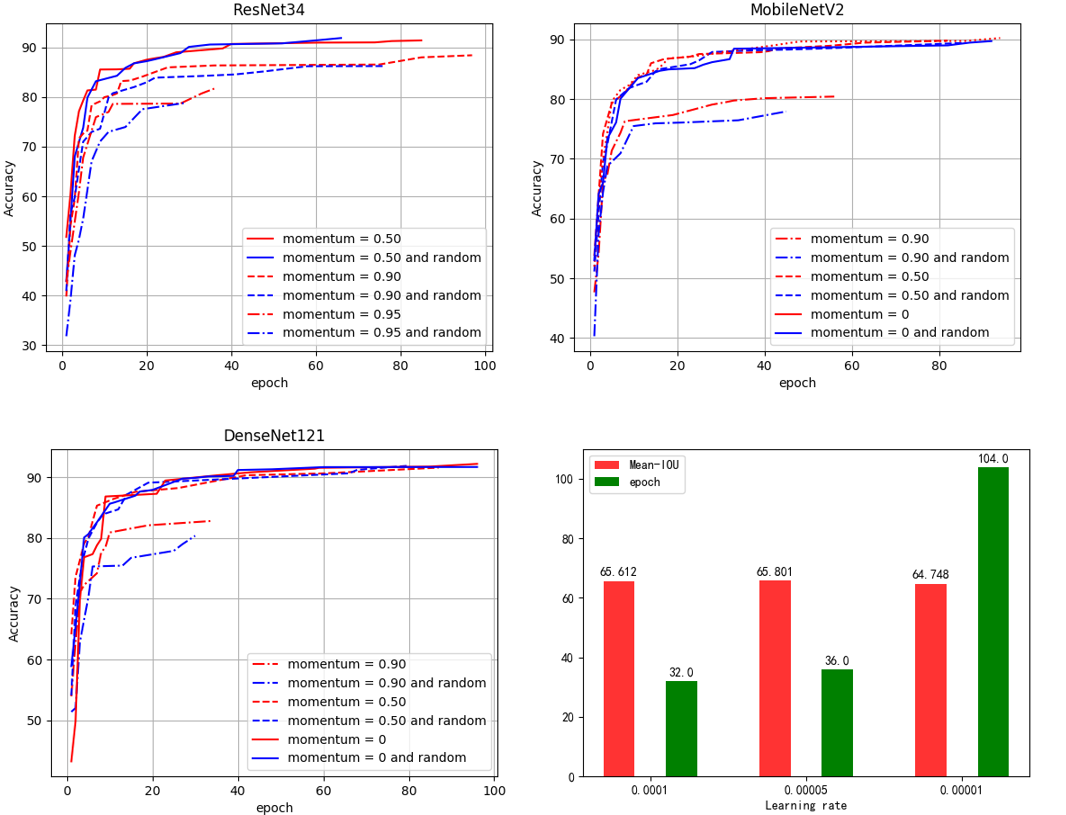}
	\caption{{\bfseries Upper left:} ResNet-34 performance on cifar 10 with different momentum. {\bfseries Upper right:} MobileNet-V2 performance on cifar 10 with different momentum. {\bfseries Bottom left:} DenseNet-121 performance on cifar 10 with different momentum. {\bfseries Bottom right:} When the loss weight is 3, the performance under different learning rates.}
	\label{fig:3}
\end{figure}
\subsection{Semantic Segmentation and Image Classification}
In Table \ref{table1}, it can be observed that when the learning rate is 0.01, the momentum is 0.95, under the combined effects of the two adverse conditions (learning rate and momentum for semantic segmentation task was too big.), both the network performance is very poor. When only reducing the learning rate to 0.001, networks with random gradients are better than the original network; however, when we fixed the learning rate (0.01) and reduce the momentum to 0.90, the result is still similar. From this we can conclude that adjusting the learning rate or momentum has the same effect in random gradients, but the learning rate can more improve the performance of the network, this is a unique feature of random gradients. When the learning rate is 0.0001 and the momentum is 0.95, the accuracy of a random gradient is slightly worse than the original gradient, this also shows that a better learning rate is more efficient than momentum.

This conclusion also applies to image classification, as shown in the left side of Figure~\ref{fig:7}, when lr is 0.1 and momentum is 0.95, both models perform poorly, when one of the conditions is changed, and the results are greatly improved.
\begin{figure}
	\centering
	\includegraphics[width=12.5cm]{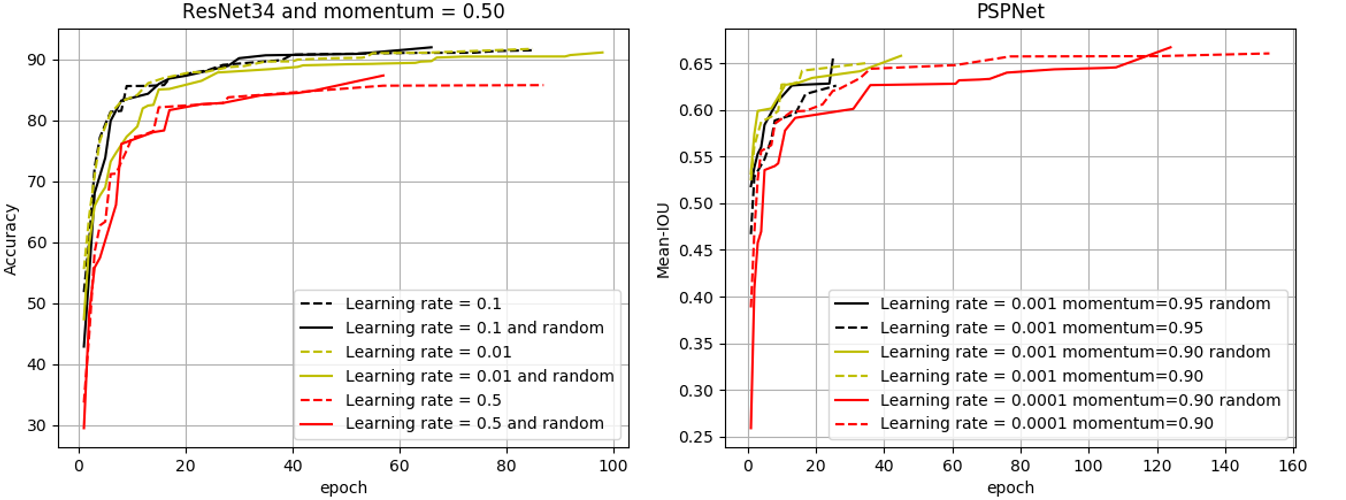}
	\caption{"random" represents a random gradient method used in the experiment. {\bf Left}: Comparison between random gradient and original gradient when change learning rate. {\bf Right}: Comparison between random gradient and original method when change momentum or learning rate.}
	\label{fig:4}
\end{figure}

In the image classification task, we selected three excellent performance networks: ResNet \cite{Authors17}, DenseNet \cite{Authors18} and MobileNetV2~\cite{Authors30}, we verify the relationship between momentum and model's accuracy. For the sake of simplicity, we fixed the initial learning rate to 0.1 and constantly adjusted the value of momentum and loss weight, it can be observed in Figure~\ref{fig:3} that the optimal momentum value for each model is not fixed. In general, setting a momentum of 0.5 in an image classification task will be a good choice. About the speed of convergence, it can be clearly seen in ResNet34 that the speed of convergence has accelerated, however, there is no obvious speed advantage in other models. In this article, we do not intend to continue to discuss in depth how ResNet network architecture relates to accelerated convergence, but we think this should be an interesting issue, because we tested a lot of models, ResNet and its variants~\cite{Authors17,Authors32,Authors33,Authors34} seem to be easier to get better results, the right side of Figure~\ref{fig:7} is our further experiment. About the accuracy rate, the improvement was not obvious in some cases, mainly because we fixed the learning rate.

In Fig~\ref{fig:4}, we put more attention on the learning rate. In the left side, we show the results in the image classification field when the momentum is 0.5, because the classification task requires a higher learning rate, the random gradient method does not perform well when the initial learning rate is small, but when the learning rate is greater than 0.1, the model can converge faster and better. In the right side, it can be seen that there is also the same result on semantic segmentation. 

It can be seen from the above that the learning rate is very important for the convergence of the model, although the model structure is different, the optimal learning rate may be significantly different, just consistent with the hyper-parameters of normal training still can have a better result, which also reduces the burden on researchers.
\subsection{GAN}
\begin{figure}[!htbp]
	\centering
	\includegraphics[width=14cm]{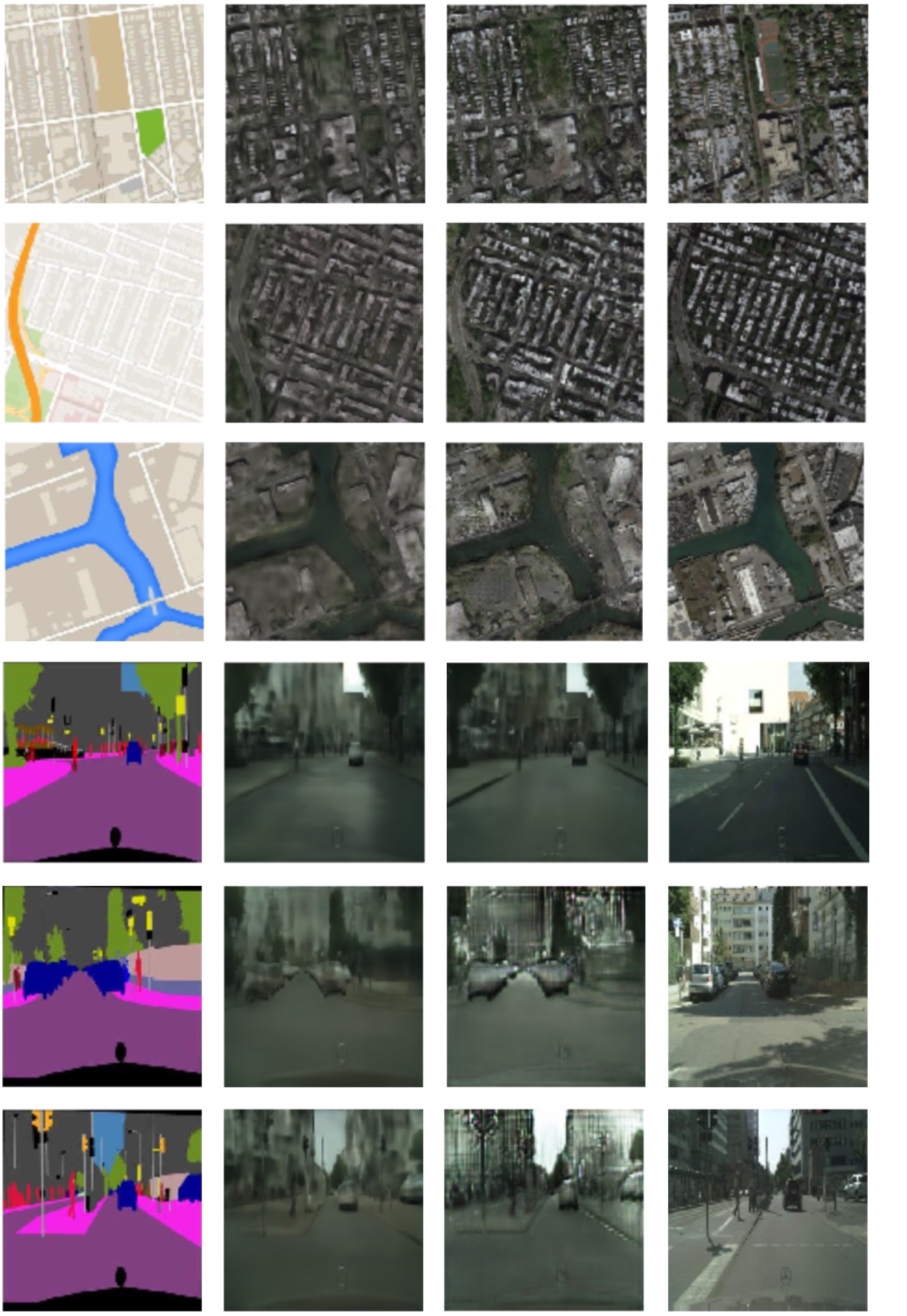}
	\caption{Visualization result on maps and cityscapes, from left to right are {\bfseries Input}, {\bfseries Original Gradient}, {\bfseries Random Gradient}, {\bfseries Ground Truth} respectively.}
	\label{fig:56}
\end{figure}

\begin{figure}[!htbp]
	\centering
	\includegraphics[width=14cm]{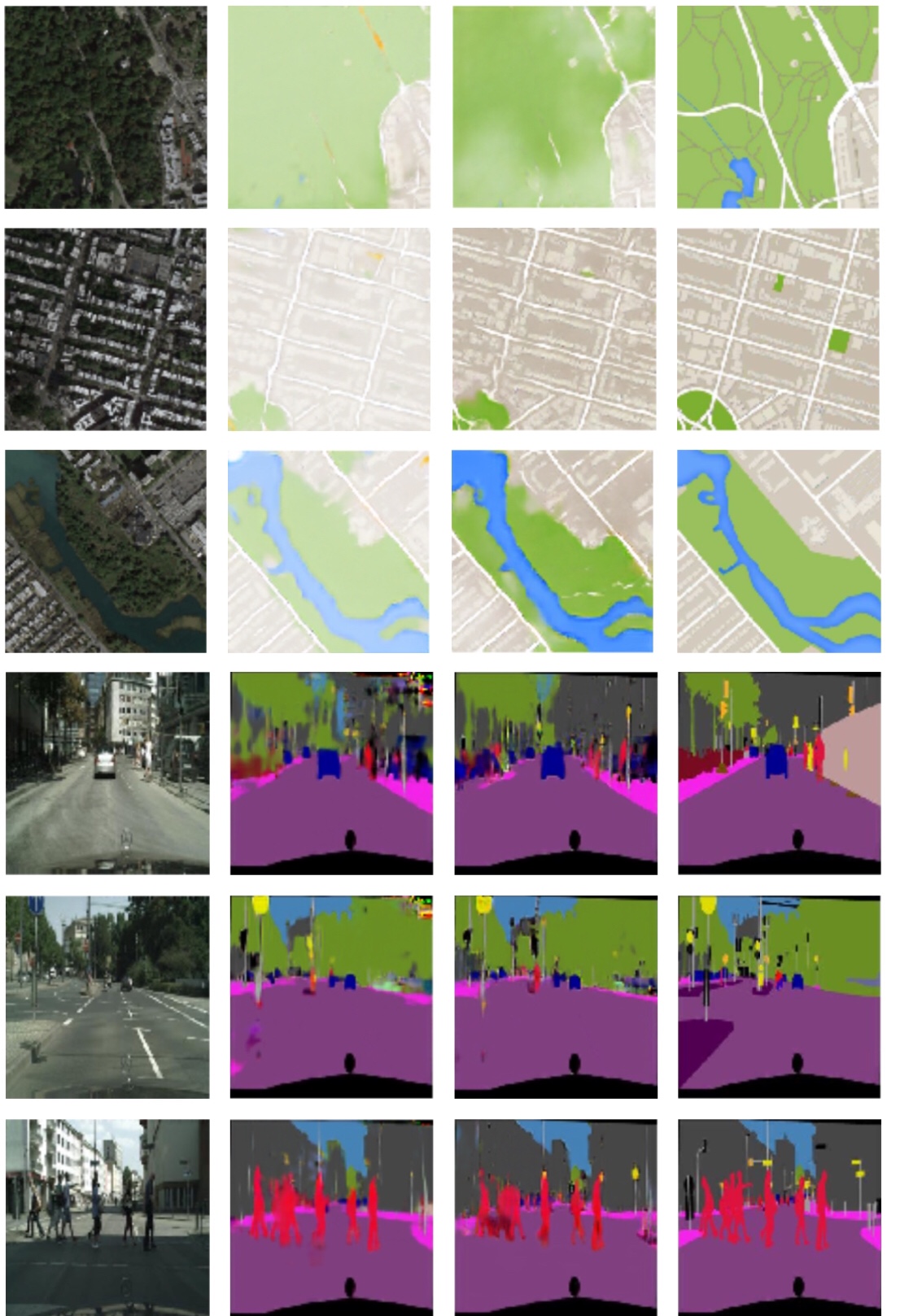}
	\caption{Visualization result on maps and cityscapes, from left to right are {\bfseries Input}, {\bfseries Original Gradient}, {\bfseries Random Gradient}, {\bfseries Ground Truth} respectively.}
	\label{fig:8}
\end{figure}

In Fig~\ref{fig:56}, \ref{fig:8}, it can be seen clearly that our method can generate clearer and more realistic images. In generating tasks, we do not test the relationship between learning rate and momentum, so we accord the method which mentioned in the paper~\cite{Authors46} to do our experiments. We apply the Adam solver~\cite{Authors04}, with learning rate 0.0002, and momentum parameters $\beta_{1}$ = 0.5, $\beta_{2}$ = 0.999, we trained the network 200 epochs, please refer to the original paper for details. Particularly noteworthy is the observation that the random gradient method can also work well on Adam, this gives the random gradient a great degree of freedom.
\section{Conclusion and Limitation}
In this paper, we presented empirical evidence for a previously phenomenon that we name random gradient. Change the gradient by applying a random weight to the loss, we are surprised that the random gradient method can perform well in many fields. It get rid of the dependence on the optimization algorithm, and using Adam in generating tasks achieves better results.

Although our method can achieve compelling results in many field, but we have not given a convincing theoretical analysis, just simply and intuitively based on the phenomenon to summarize a vague conclusion. And we have no systematic analysis of Nesterov method~\cite{Authors51}, but in some simple experimental verifications, the results are similar to the momentum method.
\section{Future Work}
\begin{flushright}\small\emph{
		~~~~~~~~~~~~~~~~~~~~~~~~~~~~~~~~~~~~~~~We can only see a short distance ahead, but we can see plenty there that needs to be done.
		-- Alan Turing}
\end{flushright}
We are pleasantly surprised to find that there is a further improvement in replacing the random number with the cyclical strategy proposed by \cite{Authors07}. It is applied to adjust the learning rate and also exhibits the nature of fast convergence under certain conditions. We will conduct more in-depth research in the future.
\clearpage
\clearpage
\clearpage

\bibliographystyle{splncs}



\bibliographystyle{plain}
\bibliography{acml18}






\end{document}